\documentclass[letterpaper, 10 pt, conference]{ieeeconf}
\IEEEoverridecommandlockouts    % to override locked commands

\usepackage{multirow}
\usepackage{lipsum}
\usepackage{amsmath}
\usepackage{amssymb}
\usepackage[ruled,vlined]{algorithm2e}
\usepackage{graphicx}
\usepackage{booktabs}
\usepackage{xcolor}
\usepackage{censor}

\title{\LARGE \bf CROSS: A Mixture-of-Experts Reinforcement Learning Framework for Generalizable Large-Scale Traffic Signal Control}

\author{Xibei Chen, Yifeng Zhang\textsuperscript{$\dag$}, Yuxiang Xiao, Mingfeng Fan, Maonan Wang, Guillaume Sartoretti
\thanks{\textsuperscript{$\dag$} Corresponding author: Yifeng Zhang.}
\thanks{Xibei Chen, Yifeng Zhang, Yuxiang Xiao, Mingfeng Fan, and Guillaume Sartoretti are with the Department of Mechanical Engineering, National University of Singapore, Singapore (E-mail: \{xibeichen, yifeng, yuxiangxiao\}@u.nus.edu, \{ming.fan, guillaume.sartoretti\}@nus.edu.sg). }
\thanks{Maonan Wang is with the Department of Mechanical and Automation Engineering, The Chinese University of Hong Kong
, Shatin, Hong Kong (E-mail: maonanwang@link.cuhk.edu.cn). }}

\begin{document}
	
\maketitle
\thispagestyle{empty}
\pagestyle{empty}

%%%%%%%%%%%%%%%%%%%%%%%%%%%%%%%%%%%%%%%%%%%%%%%%%%%%%%%%%%%%%%%%%%
\begin{abstract}
Recent advances in robotics, automation, and artificial intelligence have enabled urban traffic systems to operate with increasing autonomy towards future smart cities, powered in part by the development of adaptive traffic signal control (ATSC), which dynamically optimizes signal phases to mitigate congestion and optimize traffic.
However, achieving effective and generalizable large-scale ATSC remains a significant challenge due to the diverse intersection topologies and highly dynamic, complex traffic demand patterns across the network.
Existing RL-based methods typically use a single shared policy for all scenarios, whose limited representational capacity makes it difficult to capture diverse traffic dynamics and generalize to unseen environments.
To address these challenges, we propose CROSS, a novel Mixture-of-Experts (MoE)-based decentralized RL framework for generalizable ATSC.
We first introduce a Predictive Contrastive Clustering (PCC) module that forecasts short-term state transitions to identify latent traffic patterns, followed by clustering and contrastive learning to enhance pattern-level representation. 
We further design a Scenario-Adaptive MoE module that augments a shared policy with multiple experts, thus enabling adaptive specialization and more flexible scenario-specific strategies.
We conduct extensive experiments in the SUMO simulator on both synthetic and real-world traffic datasets.
Compared with state-of-the-art baselines, CROSS achieves superior performance and generalization through improved representation of diverse traffic scenarios.

\end{abstract}

%%%%%%%%%%%%%%%%%%%%%%%%%%%%%%%%%%%%%%%%%%%%%%%%%%%%%%%%%%%%%%%%%%
\section{INTRODUCTION}
\label{sec:introduction}

With the rapid progress of artificial intelligence and robotics, urban infrastructure is increasingly evolving into large-scale autonomous systems equipped with sensing, computation, and decision-making capabilities~\cite{zhang2024heterolight, haydari2020deep}. 
Within this context, traffic signal control has shifted from fixed timing schemes toward adaptive and learning-based strategies, giving rise to adaptive traffic signal control (ATSC).
Early ATSC systems, such as SCOOT~\cite{hunt1982scoot}, SCATS~\cite{pr1992scats} and Max-pressure control~\cite{varaiya2013max}, have shown the potential of adaptive control by utilizing loop detector data to update signal plans. 
Despite these early advancements, their rule-based nature constrains scalable and flexible autonomous decision-making under complex and rapidly changing environments.

In recent years, reinforcement learning (RL) has emerged as a highly promising approach for network-wide traffic signal control.
Early RL-based research~\cite{wei2019presslight, wei2019colight, zhang2022expression, zhang2022neighborhood, goel2023sociallight, liu2023gplight,zhang2025coordlight} primarily focused on decentralized parameter-sharing frameworks designed for homogeneous traffic networks, where intersections share identical topology and phase configurations.
Previous methods~\cite{wei2019presslight,zhang2022expression} improved traffic effectiveness by leveraging enhanced traffic states.
To further enhance cooperation among agents, researchers introduced neighborhood-communication approaches~\cite{zhang2025coordlight,zhou2024cooperative}.
However, real-world traffic networks are generally heterogeneous, with varying topologies and phase configurations.
To bridge this gap, recent studies have shifted toward developing general control frameworks as well as cross-scenario generalization frameworks.
Prior works like FRAP~\cite{zheng2019learning} and GESA~\cite{jiang2024general} address structural differences through phase competition, a standardized four-way structure and a unified state and action space. 
Recent methods have explored attention mechanisms~\cite{oroojlooy2020attendlight}, Variational Autoencoders (VAEs)~\cite{zhang2024heterolight}, meta-learning algorithms~\cite{zang2020metalight,zhu2023metavim} and diverse-training frameworks~\cite{wang2024unitsa} to improve policy adaptation. 
Building on this, Unicorn~\cite{zhang2026unicorn} integrates universal and intersection-specific traffic representations within a collaborative learning framework, enabling diverse control strategies and improved coordination among neighboring intersections.
Despite this progress, existing parameter-sharing frameworks still struggle to adapt and generalize across diverse traffic scenarios, as their representational capacity is often insufficient to handle the heterogeneity arising from diverse intersection topologies and dynamic demand patterns, thus resulting in suboptimal control performance.

To address these challenges, we propose CROSS, a MoE-based RL framework designed to enhance performance and generalizability across diverse large-scale traffic networks. 
We first adopt a General Feature Extraction (GFE) module to capture traffic dynamics and produce a unified representation that generalizes across scenarios and intersection topologies.
Building upon this backbone, we introduce a novel Predictive Contrastive Clustering (PCC) module to organize continuous traffic dynamics into a discrete set of learnable patterns. 
By anticipating short-term traffic transitions and aligning dynamic representations with representative patterns via contrastive learning, our PCC produces a compact pattern-level context.
This enables the model to capture broader traffic patterns beyond instantaneous states, thus leading to better adaptability and generalization.
To translate these pattern representations into adaptive control strategies, we design a Scenario-Adaptive MoE module.
Conditioned on the learned clustered patterns, a lightweight router selectively activates expert sub-networks.
This enables the model to activate scenario-specific policies for diverse traffic patterns, thereby circumventing the limitations of a ``one-size-fits-all'' policy that often settles for sub-optimal solutions.

We evaluate our method on synthetic and real-world datasets, and further test its performance in zero-shot transfer scenarios.
CROSS is trained on synthetic networks (e.g., \emph{Grid 5×5}) and directly evaluated on real-world datasets from \emph{Jinan} and \emph{Hangzhou}.
Results empirically show that CROSS outperforms all baselines across nearly all metrics. 
Specifically, we show that it surpasses traditional ATSC methods by effectively modeling structured traffic dynamics and adapting to varying traffic demands. 
Compared with advanced RL baselines, CROSS achieves superior control and generalization performance, benefiting from the proposed PCC module for universal pattern learning and the Scenario-Adaptive MoE architecture for expert specialization.

%%%%%%%%%%%%%%%%%%%%%%%%%%%%%%%%%%%%%%%%%%%%%%%%%%%%%%%%%%%%%%%%%%

\section{RELATED WORK}
\label{sec:related_work}

Traditional TSC methods can be broadly classified into fixed-time control and adaptive control. 
As discussed by Roess et al.~\cite{roess2004traffic}, fixed-time control operates based on a predetermined phase cycle and timing of phases, yet it struggles to adapt to complex, dynamic traffic flows. 
Alternatively, adaptive control systems such as SCOOT~\cite{hunt1982scoot} and SCATS~\cite{pr1992scats} update signal plans in response to real-time traffic conditions, leveraging traffic data collected from induction loop detectors (ILDs) to enable more responsive operation. 
Furthermore, the advanced max-pressure control~\cite{varaiya2013max} regulates intersection flow by minimizing the difference in stopped vehicle counts between upstream and downstream roads.

In recent years, RL has shown significant potential in improving network-wide traffic performance. 
The majority of existing RL-based ATSC research~\cite{wei2019presslight, wei2019colight, zhang2022expression, zhang2022neighborhood, goel2023sociallight, liu2023gplight, zhang2025coordlight} has focused on decentralized parameter-sharing frameworks designed for homogeneous networks to enhance multi-agent coordination and cooperation. 
To coordinate agents effectively, prior works such as PressLight~\cite{wei2019presslight} and Advanced-XLight~\cite{zhang2022expression} enhance their state or reward spaces by incorporating richer traffic information, including traffic pressure and advanced traffic states (ATS). 

Other approaches focus on scalable neighborhood cooperation. 
For example, NC-HDQN~\cite{zhang2022neighborhood} adjusts observations based on adjacent correlations, while SocialLight~\cite{goel2023sociallight} employs the counterfactual advantage calculation. 
Recent advancements also include CoordLight~\cite{zhang2025coordlight}'s neighbor-aware algorithm and the MICDRL~\cite{zhou2024cooperative} framework, which enhances the centralized training and decentralized execution (CTDE) paradigm via incentive communication.
Despite their effectiveness, these methods primarily focus on homogeneous traffic networks, limiting their generalizability and practicality in realistic, heterogeneous traffic environments.

To address real-world heterogeneous networks, another aspect of research focuses on cross-scenario adaptation and generalization. 
Since early independent RL methods (e.g., IA2C and MA2C~\cite{chu2019multi}) suffer from environmental instability, universal parameter-sharing frameworks have emerged to manage diverse intersections with a single shared model. 
For instance, FRAP~\cite{zheng2019learning} leverages phase competition, while GESA~\cite{jiang2024general} further enhanced FRAP by standardizing intersections into a unified four-way road structure and adopting a unified state and action space to ensure consistent input and output representations.
More recently, AttendLight~\cite{oroojlooy2020attendlight} and HeteroLight~\cite{zhang2024heterolight} introduced attention mechanisms and a VAE-based module, respectively, to extract diverse phase-conditioned representations. 
To further enhance policy adaptation across unseen scenarios, meta-learning algorithms, including MetaLight~\cite{zang2020metalight}, MetaVIM~\cite{zhu2023metavim}, and diverse-training frameworks like UniTSA~\cite{wang2024unitsa} have been explored. 
Building on these foundations, Unicorn~\cite{zhang2026unicorn} integrates a UTR module for general feature extraction and unified state-action representation, as well as an ISR module to capture diverse intersection-specific features, along with a collaborative learning to strengthen coordination among neighboring intersections.
Despite these advancements, existing universal parameter-sharing frameworks still suffer from constrained representational capacity, which limits their ability to dynamically discover latent traffic patterns and adaptively specialize strategies across diverse traffic scenarios, thus resulting in suboptimal control and generalization performance.

%%%%%%%%%%%%%%%%%%%%%%%%%%%%%%%%%%%%%%%%%%%%%%%%%%%%%%%%%%%%%%%%%%
\section{BACKGROUND}
\label{sec:background}

\begin{figure}[!t]
    \centering
    \includegraphics[width=\linewidth]{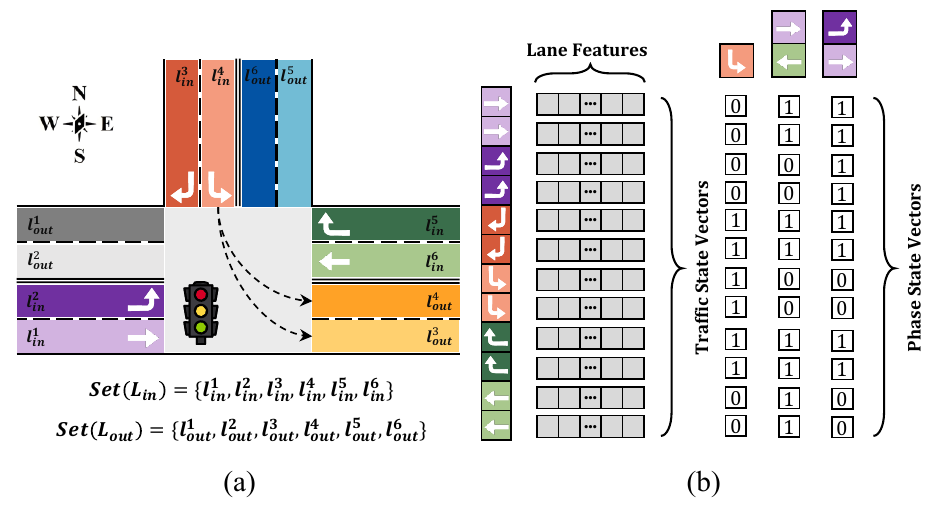}
    \caption{(a) Overview of a typical T-junction, which consists of six incoming lanes, six outgoing lanes, 12 traffic movements, and three traffic phases. (b) Traffic state and phase state vectors used in CROSS, with elements of both constructed based on ordered traffic movements.}
    \label{fig:traffic_network}
\end{figure}

\subsection{Preliminaries}

\noindent \textbf{\textit{Definition 1 (Incoming and outgoing lanes)}}: An incoming lane is a lane that approaches an intersection, while an outgoing lane is a lane that departs from an intersection.
Each road consists of multiple lanes; we denote the sets of incoming and outgoing lanes by $\mathcal{L}_{in}$ and $\mathcal{L}_{out}$, respectively.

\noindent \textbf{\textit{Definition 2 (Traffic movements)}}: A traffic movement defines a specific route through an intersection, which connects an incoming lane $l_{in}$ to an outgoing lane $l_{out}$. 
In practice, a single incoming lane may connect to multiple outgoing lanes, resulting in multiple movements.
A movement is defined as $m_{(l_{in} \to l_{out})}$ and is represented by a binary variable, where $m_{(l_{in} \to l_{out})}=1$ indicates that the movement is permitted and $m_{(l_{in} \to l_{out})}=0$ indicates that it is prohibited.

\noindent \textbf{\textit{Definition 3 (Traffic signal phases)}}: 
Each intersection maintains a predefined set of possible traffic signal phases, denoted as $\mathcal{P}$.
A specific phase $p \in \mathcal{P}$ is defined as a set of non-conflicting traffic movements that are activated simultaneously, expressed as $p = \{m=1 \mid m \in \mathcal{M}^p\}$, where $\mathcal{M}^p$ denotes the movements assigned to phase $p$. 
Accordingly, the complete set of traffic movements at an intersection, denoted by $\mathcal{M}$, is given by 
$\mathcal{M} = \bigcup_{p \in \mathcal{P}} \mathcal{M}^p$.

\noindent \textbf{\textit{Definition 4 (Traffic agents and traffic networks)}}: Traffic agents control traffic by optimizing signal phases and their timings. 
A traffic network is modeled as a multi-agent system composed of these agents. Such networks can be homogeneous, where all intersections share the same topology and phase configuration, or heterogeneous, where intersections differ in their structural layout and phase settings. 
Fig.~\ref{fig:traffic_network}(a) shows a T-junction with three incoming and three outgoing roads. 
Each road has two lanes, resulting in six incoming lanes, six outgoing lanes, and three signal phases in total.

\subsection{Multi-agent Reinforcement Learning}
In a fully decentralized setting where each intersection is controlled by an RL agent, we formulate the Multi-Agent Traffic Signal Control (MATSC) task as a Decentralized Partially Observable Markov Decision Process (Dec-POMDP)~\cite{oliehoek2016concise}. 
Formally, this Dec-POMDP is defined by the tuple $\langle\mathcal{I}, \mathcal{S}, \mathcal{A}, \mathcal{O}, \mathfrak{P}, \mathcal{R}, \gamma, \rho_{0} \rangle$, where $\mathcal{I}=\{1,2, \dots, N\}$ represents the set of agents, and $s \in \mathcal{S}$ denotes the unobservable global traffic state.
At each time step $t$, agent $i \in \mathcal{I}$ acquires a local observation $o_i^t$ through the observation function $\mathcal{O}\left(s_t, i\right)$ and selects an action $a_i^t \in \mathcal{A}_i$ based on its policy $\pi(\cdot \,|\, o_i^t)$.
These individual decisions constitute the joint action $a_t \in \mathcal{A}$. Upon executing $a_t$, each agent $i$ receives an individual reward $r_i^t$ determined by the reward function $\mathcal{R}_i(o_i^t, a_i^t)$, and the environment transitions to the next state $s^{t+1}$ according to the transition dynamics $\mathfrak{P}\left(s_{t+1} \,|\, s_t, a_t\right)$. 
Finally, $\gamma$ serves as the discount factor, and $\rho_0$ defines the initial states of distribution.
The ultimate objective is to learn an optimal joint policy $\pi^*$ that maximizes the expected discounted return across all agents: $J(\pi) = \mathbb{E}_{\tau}\left[\sum_{i=1}^{|\mathcal{I}|} \,\sum_{t=1}^{t_e} \gamma^t \, r_i^t\right]$, where $\tau=\{(o^{t},a^{t},r^{t})\}_{t=0}^{t_e}$ denotes the global trajectory with sequence length $t_e$.

\subsection{RL Agent Design}
In this study, the state, action and reward definitions for our RL agents are defined as follows: 
\subsubsection{State}
\label{state}
In ATSC problems, lane-feature vectors, which aggregate various features like queue lengths, vehicle counts, vehicle velocities, densities, and pressure, have been widely used to represent local traffic conditions in prior works~\cite{goel2023sociallight, wei2019colight, chen2020toward, zhang2022expression, oroojlooy2020attendlight, chu2019multi}.
For each time step $t$, we define the state vector for a single traffic movement $m_{(l_{in} \to l_{out})}$ at an intersection $i$ as a five-dimensional feature vector:
\begin{equation}
   S^{m}_i(t) \in \mathbb{R}^{5} =[P^{in}(t), Q^{in}(t), Q^{out}(t), N^{in}(t), N^{out}(t)],
\end{equation}
\noindent where $P^{in}(t)$ is the current movement activation status, $Q^{in}(t)$ and $Q^{out}(t)$ respectively represent the number of stopped vehicles (queue length) on the incoming and outgoing lanes,  while $N^{in}(t)$ and $N^{out}(t)$ indicate the count of moving vehicles. 
These features are typically collected via intersection cameras. 
Therefore, the local \textit{traffic state vector} for a single intersection $i$ can be represented as:
\begin{equation}
\label{state_function}
S_i^t = S_i(t) \in \mathbb{R}^{|\mathcal{M}_i| \times 5}=[S_i^m(t) \mid m \in \mathcal{M}_i],    
\end{equation}
\noindent which includes the states of all available traffic movements.
We also define the time-invariant \textit{phase state vector} that shows the movement activation for a given phase $p \in \mathcal{P}_i$:
\begin{equation}
    G^p_i \in \mathbb{R}^{|\mathcal{M}_i|} =[1 \text { if } m \in \mathcal{M}_{i}^{p} \text { else } 0 \mid m \in \mathcal{M}_i],
\end{equation}
\noindent where $\mathcal{M}_i$ is the set of all traffic movements and $\mathcal{M}_{i}^{p}$ is the subset activated by phase  $p$. 
For the entire phase set $\mathcal{P}_i$, the comprehensive phase state is $G_i$, which is represented as: 
\begin{equation}
G_i \in \mathbb{R}^{|\mathcal{P}_i| \times |\mathcal{M}_i|} = [G^p_i \mid p \, \in \, \mathcal{P}_i].
\end{equation}
A detailed example of a T-junction's traffic state and phase state vectors is shown in Fig.~\ref{fig:traffic_network}(b).
Furthermore, the time-invariant \textit{topology vector} for an intersection is defined by: 
\begin{equation}
I_i = [T_{\text{tl}}, L^{in}, V_{\text{max}}^{in}, N_{\text{l}}^{in}, N_{\text{m}}^{in}, L^{out}, V_{\text{max}}^{out}, N_{\text{l}}^{out}],
\end{equation}
where $T_{\text{tl}}$ is a one-hot vector of the intersection type and phase configuration. 
$L^{in}$ ,$V_{\text{max}}^{in}$, $N_{\text{l}}^{in}$ and $N_{\text{m}}^{in}$ represent the average lane length, maximum speed limit, lane count, and total movement count for incoming roads, respectively. 
Corresponding metrics are included for the outgoing roads: $L^{out}$, $V_{\text{max}}^{out}$, and $N_{\text{l}}^{out}$ represent the average lane length, maximum speed, and the number of lanes, respectively. 
%======================Figure============================
\begin{figure*}[ht]
    \centering
    \includegraphics[width=0.8\textwidth,trim=2cm 0cm 2cm 0,
    ]{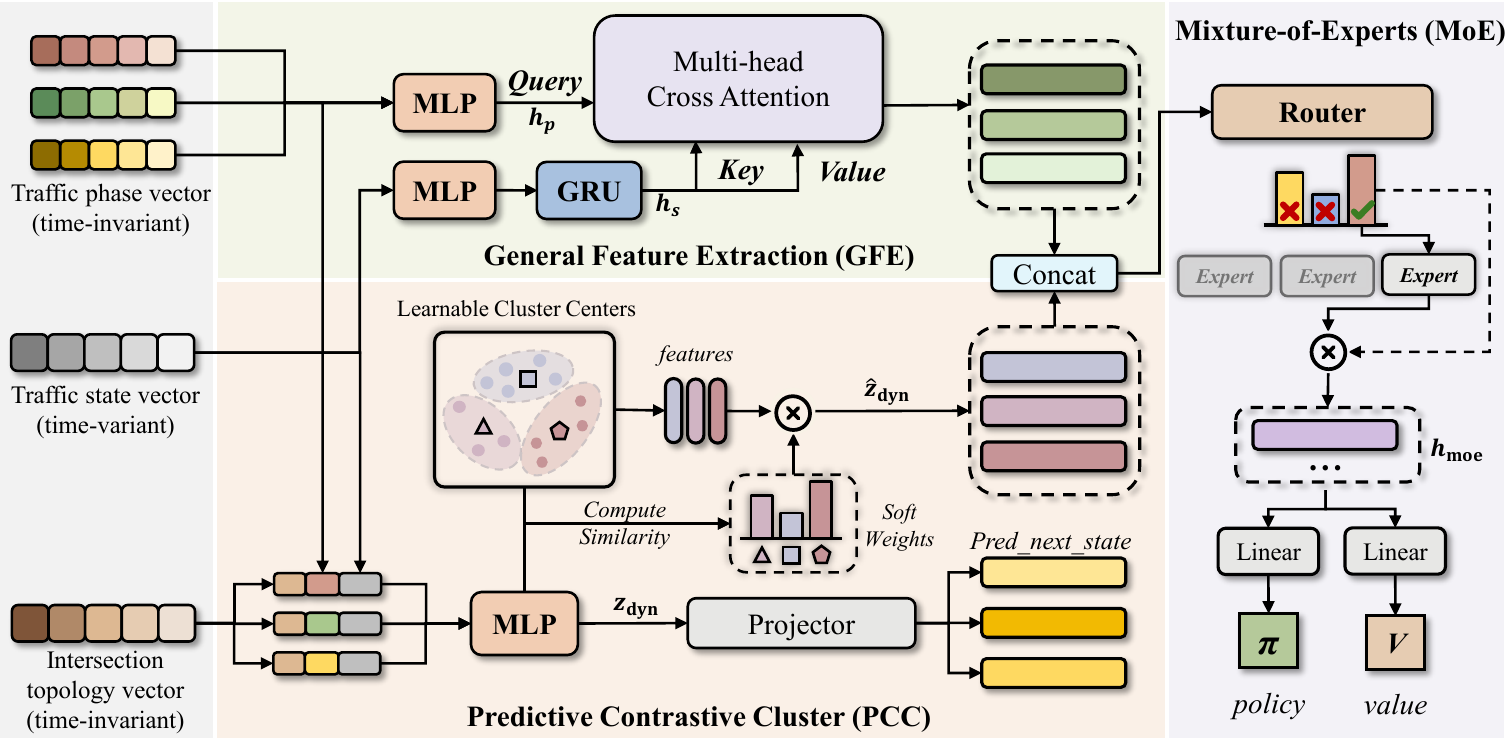}
    \caption{Architecture of the proposed CROSS framework, which includes a GFE module to extract rich traffic dynamics and form a unified representation for heterogeneous intersections, a PCC module to predict traffic dynamics and match them with learnable cluster centers for generalizable pattern representations, and a Scenario-Adaptive MoE module that uses these representations as routing context to compute weights and generate the traffic policy and value functions.}
    \label{fig:method}
\end{figure*}

\subsubsection{Action}
In this study, the action space for each agent consists of a finite set of collision-free traffic phases. 
Agents simultaneously select and implement a phase from these sets for a predetermined duration.  
Unlike traditional signal control, our agents are not bound to a fixed cycle, an approach widely adopted in recent ATSC works~\cite{wei2019presslight, wei2019colight, chen2020toward, chu2019multi, goel2023sociallight}. 
This flexibility allows the system to skip unnecessary phases, thus improving control adaptability.

\subsubsection{Reward}
The reward for each agent is designed to minimize congestion by penalizing long queues. 
Specifically, the reward is defined as the negative sum of queue lengths measured by induction loop detectors (with an effective range of 50 meters) located on the incoming lanes of the intersection. 
This is formulated as:
$r(t)=-\left(\sum_{l_{in} \in \mathcal{L}_{in}} Q^{l_{in}}(t)\right)$.

%%%%%%%%%%%%%%%%%%%%%%%%%%%%%%%%%%%%%%%%%%%%%%%%%%%%%%%%%%%%%%%%%%
\section{CROSS}
\label{sec:method}
An overview of the proposed CROSS framework is illustrated in Fig.~\ref{fig:method}. 
Specifically, it comprises three key components: 
(1) a GFE module that extracts rich traffic dynamics and produces unified representation, laying the foundation for scalable deployment in diverse intersections;
(2) a PCC module that predicts traffic dynamics and matches them with learnable clustering centers to obtain a generalizable pattern-level representation;
(3) a Scenario-Adaptive MoE that leverages these representations as routing contexts.
The MoE then generates routing weights to explicitly modulate the unified representation provided by the GFE module.

\subsection{General Feature Extraction (GFE)}
To build a unified and expressive traffic representation, we adopt the General Feature Extraction (GFE) module~\cite{zhang2024heterolight} as the representation backbone of CROSS. 
This module systematically extracts traffic dynamics by encoding temporal evolution and phase-specific structural information, providing a scalable basis for deployment in real-world heterogeneous intersections.
For each agent, the traffic state vector is encoded into a latent representation $h_s$ with temporal dependencies captured by a Gated Recurrent Unit (GRU)~\cite{chung2014empirical}, while phase representations $h_p$ provided by an MLP encode the structural characteristics of available phases. 
A multi-head cross-attention mechanism then integrates state and phase information to produce phase-aware representations $h_{sp} \in \mathbb{R}^{|\mathcal{P}_i| \times d}$, enabling effective representation extraction across heterogeneous intersections.
The output of the GFE module is thus a general representation that encodes both temporal traffic evolution and phase-specific structural information. 

\subsection{Predictive Contrastive Clustering (PCC)}
To effectively categorize diverse traffic dynamics into representative patterns for scenario-adaptive policy routing, we propose the PCC module. 
Specifically, for agent $i$ under traffic phase $p$ at time step $t$, the input to the PCC module is defined as the concatenation of three complementary information sources: $x_i^p(t) =[S_i^t, G_i^p, I_i]$,
where $S_i^t$ is the current traffic state vector encoding movement-level representations, $G_i^p$ is the phase state vector indicating the activation status of traffic movements, and $I_i$ is the intersection topology vector.
We first obtain the predictive dynamics representation $z_{\text{dyn}} \in \mathbb{R}^{d_z}$ by passing $x_i^p(t)$ through a two-layer MLP with GLU, where $d_z$ denotes the hidden dimension.
After that, a linear projection head maps it to a predicted next-step state $\hat{S}_i^{t+1} = W_{\text{proj}} \, z_{\text{dyn}}$, which is supervised by the ground truth future observation via a mean squared error (MSE) loss $L_{\text{pred}}$.
By jointly conditioning on these vectors, PCC is able to perceive not only \emph{what is happening} (traffic dynamics) but also \emph{where it is happening} (intersection geometry) and \emph{under which control} (phase configuration), thereby producing a spatially aware and action-conditioned representation $z_\text{dyn}$.

We map the continuous dynamic representations onto a discrete set of clustering centers, and thereby enable PCC to provide a compact and interpretable scene context for downstream expert selection.
Specifically, this module maps the dynamics representation $z_{\text{dyn}}$ onto a set of learnable clustering center vectors 
$\mathcal{C} = \{c_k\}_{k=1}^{K}$, each representing a characteristic traffic pattern. 
Given $z_{\text{dyn}}$, PCC first projects it into the cluster space and computes the cosine similarity with each clustering center: $\text{sim}(z_{\text{dyn}}, c_k)=
\frac{
    \phi(z_{\text{dyn}})^\top c_k
}{
    \lVert \phi(z_{\text{dyn}}) \rVert_2
    \lVert c_k \rVert_2
},$
where $\phi(\cdot)$ denotes a projection head consisting of a one-layer MLP followed by Layer Normalization. 
A softmax function with temperature $\tau_k$ is then applied to obtain a soft assignment weight $w_k$ for each center:
$w_k = \frac{\exp(\text{sim}(z_{\text{dyn}}, c_k) / \tau_k)}{\sum_{j=1}^{K} \exp(\text{sim}(z_{\text{dyn}}, c_j) / \tau_k)}.$
The final quantized representation is computed as a weighted sum of clustering center vectors $\hat{z}_{\text{dyn}} = \sum_{k=1}^{K} w_k \, c_k$.

To optimize PCC, we adopt an InfoNCE-style contrastive loss~\cite{oord2019representationlearningcontrastivepredictive}, because its mutual-information maximization and softmax-based structure align naturally with expert selection.
Denoting $s_k = \text{sim}(z_{\text{dyn}},\, c_k) / {\tau_c}$ as the temperature-scaled similarity between a sample and the $k$-th cluster center, where $\tau_c$ is a temperature parameter.
Therefore, the contrastive loss is formulated as follows:
\begin{equation}
L_{\text{cont}} = -\mathbb{E} \left[ \sum_{k=1}^{K} w_k \log \frac{\exp(s_k)}{\sum_{j=1}^{K} \exp(s_j)} \right],
\end{equation}
where $w_k$ serves as the soft-target distribution representing the relevance of the $k$-th center. 
This formulation encourages $z_{\text{dyn}}$ to be pulled toward the cluster centers with high assignment weights while being pushed away from others, thereby forming a discriminative representation space.

To prevent mode collapse where only a subset of clusters is actively utilized, we also add a diversity regularization loss based on the entropy of the batch-averaged assignment:
\begin{equation}
L_{\text{div}} = 1 - \frac{H(\bar{\mathbf{w}})}{\log K}, 
\quad \bar{\mathbf{w}} = \frac{1}{B}\sum_{i=1}^{B} \mathbf{w}^{(i)},
\end{equation}
where $H(\cdot)$ denotes Shannon entropy, $B$ is the batch size, and $\mathbf{w}^{(i)}$ is the soft assignment vector of the $i$-th sample. 
This loss reaches zero when all cluster centers are uniformly utilized and increases towards one when assignments collapse onto a single center.
Thus, the total PCC loss is:
\begin{equation}
L_{\text{PCC}} =
L_{\text{cont}} + \lambda_{\text{div}} L_{\text{div}}.
\end{equation}
where $\lambda_{\text{div}}$ is a weighting coefficient.
The resulting quantized representation $\hat{z}_{\text{dyn}}$ serves as the context pattern for the downstream MoE router, enabling scenario-adaptive expert activation based on the identified traffic pattern.

\subsection{Scenario-Adaptive Mixture-of-Experts}

The Scenario-Adaptive MoE in CROSS consists of a router and a set of $N_E$ expert subnetworks $\{E_m(\cdot)\}_{m=1}^{N_E}$. 
Each expert is implemented as a lightweight MLP that transforms the spatial-temporal representation $h_{sp}$ from GFE into a specialized representation $h_m = E_m(h_{sp})$. 
The router determines which experts should be activated for a given traffic state, enabling adaptive policy specialization while reducing cross-scenario conflicts during joint training.

Concretely, the router takes as input the concatenation of the state representation $h_{sp}$ and the clustered dynamics representation $\hat{z}_{\text{dyn}}$, and produces routing weights $w_r$: $w_r = \text{MLP}_{\text{router}}([h_{sp}, \hat{z}_{\text{dyn}}]) / \tau_r,$
where $\tau_r$ is a temperature hyperparameter.
A hard top-$k$ gating mechanism is applied to sparsely activate the most relevant experts according to $w_r$. 
Let $\alpha_m$ denote the normalized gating weight of expert $m$, where only the top-$k$ experts receive non-zero weights. 
The final aggregated representation $h_{\text{moe}}$ is computed as: $h_{\text{moe}}=\sum_{m=1}^{N_E} \alpha_m \, E_m(h_{sp}).$
This design allows our method to selectively activate relevant experts according to the identified traffic pattern, ensuring specialized policy adaptation for different scenarios while promoting stable training.

To regularize the routing behavior, we introduce two auxiliary losses. 
First, a load balancing loss $L_{\text{lb}}$ penalizes imbalanced expert utilization by minimizing the KL divergence between the empirical expert usage distribution and a uniform prior, which is calculated as follows:
\begin{equation}
L_{\text{lb}} = \sum_{m=1}^{N_E} \hat{f}_m 
\log \frac{\hat{f}_m}{1/N_E},
\end{equation}
where $\hat{f}_m = \frac{\sum_{i=1}^{B} \alpha_m^{(i)}}{\sum_{m'}\sum_{i} \alpha_{m'}^{(i)}}$ is the normalized usage frequency of expert $m$ across the batch.
This loss encourages all experts to receive a comparable share of routing traffic, preventing expert starvation. 
Second, a selection entropy loss $L_{\text{se}}$ encourages confident, decisive routing by minimizing the entropy of each sample's routing 
weight distribution:
\begin{equation}
L_{\text{se}} = -\frac{1}{B}\sum_{i=1}^{B}
\sum_{m=1}^{N_E} \alpha_m^{(i)} 
\log \alpha_m^{(i)}.
\end{equation}
Intuitively, $L_{\text{lb}}$ and $L_{\text{se}}$ serve complementary roles: 
the former promotes macro-level diversity across the expert pool, while the latter enforces micro-level certainty for each routing decision. 
The combined MoE loss is calculated as:
\begin{equation}
L_{\text{MoE}} = \lambda_{\text{lb}}\,L_{\text{lb}} + \lambda_{\text{se}}\,L_{\text{se}}.    
\end{equation}
Here $\lambda_{\text{lb}}$ and $\lambda_{\text{se}}$ are balancing coefficients that control the relative strength of MoE load balancing and routing certainty regularization terms.
After obtaining the aggregated representation $h_{\text{moe}}$, we feed it into two separate linear layers to compute the policy and value functions.

Notably, the experts are not pre-trained; instead, they are jointly optimized with the lightweight router in an end-to-end, fully differentiable manner during online training.

\subsection{Policy Optimization}
In this work, we employ Proximal Policy Optimization (PPO)~\cite{schulman2017proximal} to optimize the policy function $\pi_\theta$ with parameters $\theta$ and the value function $V_\Phi$ with parameters $\Phi$, given its widespread adoption and strong empirical performance in terms of training stability and efficiency.
Both actor and critic networks share parameters across agents, which improves training efficiency and enables scalable signal control.

For each agent $i$, the policy loss is defined as:
\begin{equation}
L^{a}_i(\theta) = -\mathbb{E}_t\left[
\min\left(
\kappa_i^t(\theta)\,\hat{A}_i^t,\;
\text{clip}\!\left(\kappa_i^t(\theta),\,1{-}\epsilon,\,1{+}\epsilon\right)\hat{A}_i^t
\right)
\right],
\end{equation}
where $\kappa(\theta)$ denotes the probability ratio between the updated and behavior policies, $\hat{A}_i^t$ is the advantage estimate computed via Generalized Advantage Estimation (GAE)~\cite{schulman2015high}, and $\epsilon$ denotes the clipping threshold that limits large policy updates.

The value loss is defined as the temporal-difference error:
\begin{equation}
L^c_i(\Phi)=\mathbb{E}_t\left[\left(r^t_i + \gamma \, V_{\Phi, i}^{t+1} - V_{\Phi,i}^{t} \right)^2\right],
\end{equation}
In addition, an entropy loss $L^{H}_i(\theta)$ is added to encourage exploration and reduce premature convergence.

The overall training loss integrates the standard RL losses with the auxiliary losses introduced in our framework, including the predictive loss $L_{\text{pred}}$, the clustering loss $L_{\text{PCC}}$, and the MoE regularization loss $L_{\text{MoE}}$. 
Thus, the loss for optimizing the actor network is formulated as:
\begin{equation}
L_{\text{a}}(\theta) =
\frac{1}{N}\sum_{i=1}^{N}
\left(
L^{a}_i
-
\lambda_e\,L^{H}_i
\right)
+
\lambda_p\,L_{\text{pred}}^{a}
+
\lambda_c\,L_{\text{PCC}}^{a}
+
L_{\text{MoE}}^{a},
\end{equation}
while the loss for optimizing the critic network is defined as:
\begin{equation}
L_{\text{c}}(\Phi) =
\frac{1}{N}\sum_{i=1}^{N}
\lambda_v\,L^{c}_i
+
\lambda_p\,L_{\text{pred}}^{c}
+
\lambda_c\,L_{\text{PCC}}^{c}
+
L_{\text{MoE}}^{c}.
\end{equation}

Here, $\lambda_v$, $\lambda_e$, $\lambda_p$, and $\lambda_c$ are balancing coefficients for the respective loss components. 
The superscripts $a$ and $c$ indicate that the auxiliary modules are instantiated separately in the actor and critic networks, with gradients propagated independently.
Both networks are optimized using Adam~\cite{kingma2017adammethodstochasticoptimization} with separate parameter updates. 
To enable efficient batch training across heterogeneous intersections, traffic state and phase vectors are padded to the maximum dimensionality ($|\mathcal{M}|_{\max}{=}36$, $|\mathcal{P}|_{\max}{=}8$) within the network. 
Padding entries are masked during forward computation and loss evaluation to ensure that they do not influence optimization.
%%%%%%%%%%%%%%%%%%%%%%%%%%%%%%%%%%%%%%%%%%%%%%%%%%%%%%%%%%%%%%%%%%
\section{EXPERIMENTAL RESULTS}
\label{sec:experimental_results}
\subsection{Traffic Datasets}

We conduct experiments on three synthetic and five real-world traffic datasets.
The synthetic traffic datasets consist of \textit{Grid 4×4}~\cite{ault2021reinforcement}, \textit{Arterial 4×4}~\cite{ault2021reinforcement}, and \textit{Grid 5×5}~\cite{chu2019multi}. 
The real-world traffic datasets~\cite{wei2019colight} include $Jinan_{(1)}$, $Jinan_{(2)}$, and $Jinan_{(3)}$ from Jinan City, China, as well as $Hangzhou_{(1)}$ and $Hangzhou_{(2)}$ from Hangzhou City, China. 
In detail, \textit{Grid 4$\times$4} and \textit{Arterial 4$\times$4} each include 16 intersections, while \textit{Grid 5$\times$5} includes 25 intersections. 
These synthetic networks are regular and homogeneous in structure, with traffic demand generated according to the predefined patterns introduced in their original work.
For the real-world datasets, the $Jinan$ and $Hangzhou$ networks contain 12 and 16 intersections, respectively, with each traffic flow dataset collected from different time periods in real-world traffic. In contrast, real-world networks are structurally heterogeneous and exhibit complex realistic traffic demand.
Table~\ref{tab:selected_datasets} summarizes the statistical properties of all datasets, including total traffic volume and arrival rate distributions. 

\begin{table}[t]
\centering
\caption{Traffic demand specifications of the experimental datasets.}
\label{tab:selected_datasets}
\begin{tabular}{c c cccc}  
\toprule
\multirow{2}{*}{\textbf{Traffic Dataset}} 
& \multirow{2}{*}{\textbf{Volume (veh)}} 
& \multicolumn{4}{c}{\textbf{Arrival Rate (veh/min)}} \\
\cmidrule(lr){3-6}
& & \textbf{Mean} & \textbf{Std.} & \textbf{Max} & \textbf{Min} \\
\midrule
Grid $4\times4$     & 1473.00 & 24.55  & 13.89 & 72.00  & 6.00 \\
Arterial $4\times4$ & 2484.00 & 41.40  & 24.69 & 88.00  & 10.00 \\
Grid $5\times5$     & 7296.00 & 121.60 & 121.94 & 752.00 & 32.00 \\
\midrule
$Jinan_{(1)}$ & 6295.00  & 104.92 & 19.79 & 136.00 & 50.00 \\
$Jinan_{(2)}$ & 4365.00  & 72.75  & 15.15 & 101.00 & 43.00 \\
$Jinan_{(3)}$ & 5494.00  & 91.57  & 9.51  & 111.00 & 69.00 \\
$Hangzhou_{(1)}$ & 2983.00  & 49.72  & 8.24  & 67.00  & 40.00 \\
$Hangzhou_{(2)}$ & 6984.00  & 116.40 & 63.72 & 230.00 & 39.00 \\
\bottomrule
\end{tabular}
\end{table}

\subsection{Baselines and Metrics}

We compare our CROSS framework with both conventional and advanced RL-based ATSC methods. 
The \textbf{conventional approaches} include Fixed-Time and Max-Pressure~\cite{varaiya2013max}, which operate based on predetermined phase cycle or rules. 
For \textbf{RL-based methods}, we include GESA~\cite{jiang2024general}, which uses a unified state--action space and multi-scenario joint training to improve generalization across diverse networks,
as well as Unicorn~\cite{zhang2026unicorn}, which adopts a universal state-action representation, and leverages UTR and ISR modules to enable adaptable and universal policy learning for heterogeneous traffic networks.
Notably, GESA and Unicorn are state-of-the-art multi-scenario co-training frameworks; therefore, we focus on comparisons with these representative methods rather than exhaustively reproducing all baselines reported in their original studies.

Following common practice in prior work~\cite{chu2019multi, zhang2026unicorn}, we evaluate performance using multiple key network-level metrics. 
Specifically, we report \textit{average queue length}, \textit{average speed}, \textit{trip completion rate}, \textit{average trip time}, \textit{average trip delay}, and \textit{average trip duration}.
Among these, \textit{average trip duration} offers a more comprehensive assessment, as it accounts for both completed trips and delays from vehicles that have not yet departed or failed to finish.

\subsection{Experiment Settings}
In our experiments, we consider two evaluation settings. 
First, CROSS and the RL-based baseline methods are co-trained across multiple synthetic traffic scenarios and then directly evaluated on these synthetic datasets to assess cross-scenario control performance.
Second, we evaluate the zero-shot performance of these methods on real-world datasets after training them solely on synthetic data, which mirrors practical sim-to-real deployment under real-world data scarcity. 
Since conventional methods operate based on fixed phase cycles or predetermined rules, they require no training and are evaluated directly in both settings.

We conduct our experiments on the open-source traffic simulator SUMO~\cite{SUMO2018}. 
Following RESCO~\cite{ault2021reinforcement} setting, we adopt a 10-second green phase duration and a 3-second yellow transition.
Each simulation episode spans 3600 seconds.
The discount factor is set to 0.95 with a GAE parameter of 0.98. 
The actor and critic networks are optimized using separate Adam optimizers with learning rates of $1\times 10^{-4}$ and $2\times 10^{-4}$, respectively. 
The backbone MLP hidden dimension is 128. 
For the PPO algorithm, we use a clip ratio of 0.2 and perform 6 update epochs per training iteration. Gradient norms are clipped at 10.
For the CROSS modules, the PCC clustering hidden dimension is set to 64, and it employs $K{=}6$ learnable centers with an assignment temperature $\tau_k{=}0.1$ and a contrastive temperature $\tau_c{=}0.1$. 
The MoE architecture consists of $6$ experts with top-2 routing. 
The loss balancing coefficients are set as: $\lambda_v{=}0.5$, $\lambda_{\text{e}}{=}0.01$, $\lambda_{\text{p}}{=}0.05$, $\lambda_{\text{c}}{=}0.1$, $\lambda_{\text{lb}}{=}0.001$, and $\lambda_{\text{se}}{=}0.0001$. 
All methods are trained for approximately 3000 episodes to ensure convergence under the multi-scenario co-training regime.
Our code implementation and datasets will be released upon acceptance of the paper.

\begin{table*}[t]
\centering
\caption{Performance comparison across synthetic training datasets and real-world evaluation scenarios.$\downarrow$ denotes lower is better, $\uparrow$ denotes higher is better. Best values are \textbf{bold}, second-best are \underline{underlined}.}
\label{tab:synthetic_results}
\renewcommand{\arraystretch}{1.3}
\resizebox{\textwidth}{!}{
\begin{tabular}{c|ccccc|ccccc}
\hline
\multirow{3}{*}{\textbf{Method}} 
& \multicolumn{5}{c|}{\textbf{Synthetic Dataset Evaluation}} 
& \multicolumn{5}{c}{\textbf{Real-World Dataset Evaluation (Zero-shot)}} \\ \cline{2-11}
& \begin{tabular}[c]{@{}c@{}}Queue Length $\downarrow$ \\ (veh)\end{tabular} 
& \begin{tabular}[c]{@{}c@{}}Speed $\uparrow$ \\ (m/s)\end{tabular} 
& \begin{tabular}[c]{@{}c@{}}Trip Comp. Rate $\uparrow$ \\ (veh/s)\end{tabular} 
& \begin{tabular}[c]{@{}c@{}}Trip Time $\downarrow$ \\ (s)\end{tabular} 
& \begin{tabular}[c]{@{}c@{}}Trip Delay $\downarrow$ \\ (s)\end{tabular} 
& \begin{tabular}[c]{@{}c@{}}Queue Length $\downarrow$ \\ (veh)\end{tabular} 
& \begin{tabular}[c]{@{}c@{}}Speed $\uparrow$ \\ (m/s)\end{tabular} 
& \begin{tabular}[c]{@{}c@{}}Trip Comp. Rate $\uparrow$ \\ (veh/s)\end{tabular} 
& \begin{tabular}[c]{@{}c@{}}Trip Time $\downarrow$ \\ (s)\end{tabular} 
& \begin{tabular}[c]{@{}c@{}}Trip Delay $\downarrow$ \\ (s)\end{tabular} \\ 
\hline

\multicolumn{6}{c|}{\textbf{Grid $4\times4$ (Easy, Synthetic)}} 
& \multicolumn{5}{c}{\textbf{$Jinan_{(1)}$ (Hard, Real-World)}} \\
\hline
Fixed-Time     & 0.12(0.09) & 7.88(1.28)  & 0.40(0.41) & 194.63(76.87)  & 54.96(39.15)   & 2.02(0.93) & 5.23(1.35) & 1.36(0.65) & 461.97(368.94) & 170.23(249.98) \\
Max-Pressure   & 0.07(0.06) & 8.82(0.66)  & 0.40(0.35) & 174.41(63.07)  & 33.18(23.61)   & \textbf{0.33(0.14)} & \textbf{8.68(0.37)} & \textbf{1.60(0.62)} & \textbf{288.10(142.99)} & \textbf{25.95(28.56)} \\
\hline
GESA           & \textbf{0.04(0.04)} & \textbf{10.08(0.73)} & \textbf{0.42(0.37)} & \textbf{154.03(53.46)} & \textbf{19.12(19.70)} & 0.94(0.36) & 7.33(0.70) & 1.56(0.60) & 341.84(200.87) & 76.53(97.96) \\
Unicorn        & 0.05(0.04) & 9.55(0.59) & 0.40(0.35) & 160.38(53.93) & 22.45(20.89) & 3.81(1.98) & 4.04(1.65) & 1.29(0.51) & 593.21(513.83) & 313.28(389.58) \\
CROSS          & \underline{0.05(0.04)} & \underline{9.61(0.56)}  & \underline{0.40(0.35)} & \underline{158.63(54.21)}  & \underline{20.88(19.36)} & \underline{0.65(0.22)} & \underline{7.94(0.41)} & \underline{1.59(0.63)} & \underline{314.27(169.16)} & \underline{53.03(66.23)} \\
\hline

\multicolumn{6}{c|}{\textbf{Arterial $4\times4$ (Medium, Synthetic)}} 
& \multicolumn{5}{c}{\textbf{$Jinan_{(2)}$ (Medium, Real-World)}} \\
\hline
Fixed-Time     & 2.62(1.29) & 1.45(1.15)  & 0.29(0.45) & 839.70(529.19)   & 597.61(424.47)   & 0.77(0.25) & 6.76(0.76) & 1.12(0.55) & 377.90(244.70) & 90.51(118.79) \\
Max-Pressure   & 1.11(0.59) & 3.42(1.65)  & \textbf{0.56(0.30)} & 380.29(225.67)   & \textbf{117.03(116.25)} & \textbf{0.17(0.07)} & \textbf{9.03(0.30)} & \textbf{1.15(0.47)} & \textbf{284.63(138.34)} & \textbf{18.87(19.10)} \\
\hline
GESA           & 3.44(1.76) & 1.17(2.24)  & 0.06(0.12) & 1819.84(1128.22) & 1721.50(1145.27) & 0.40(0.14) & 8.28(0.42) & \underline{1.15(0.47)} & 311.68(161.52) & 46.51(59.14) \\
Unicorn        & \underline{0.98(0.45)} & \underline{3.75(1.79)} & 0.53(0.28) & \textbf{289.98(276.42)} & \underline{130.46(238.56)} & 1.84(0.60) & 5.04(1.167) & 1.06(0.42) & 504.28(416.48) & 218.11(289.53) \\
CROSS          & \textbf{0.95(0.45)} & \textbf{3.76(1.71)} & \underline{0.54(0.29)} & \underline{312.08(350.87)} & 152.08(331.14) & \underline{0.38(0.12)} & \underline{8.29(0.33)} & 1.14(0.47) & \underline{309.88(159.79)} & \underline{44.74(55.42)} \\
\hline

\multicolumn{6}{c|}{\textbf{Grid $5\times5$ (Hard, Synthetic)}} 
& \multicolumn{5}{c}{\textbf{$Jinan_{(3)}$ (Hard, Real-World)}} \\
\hline
Fixed-Time     & 2.70(1.65) & 3.02(2.11)  & 0.84(0.41) & 710.59(542.07)   & 403.80(466.43)   & 1.44(0.59) & 5.82(1.05) & 1.27(0.59) & 418.73(333.50) & 139.39(235.26) \\
Max-Pressure   & 3.25(1.66) & 2.01(1.58)  & 0.61(0.54) & 817.36(661.51)   & 569.91(554.88)   & \textbf{0.23(0.07)} & \textbf{8.88(0.29)} & \textbf{1.40(0.52)} & \textbf{277.47(136.62)} & \textbf{20.68(21.53)} \\
\hline
GESA           & 6.80(1.81) & 0.92(0.63)  & 0.66(0.38) & 1129.55(1351.83) & 1012.28(1334.17) & 0.67(0.19) & 7.66(0.49) & 1.38(0.52) & 320.96(176.28) & 62.35(84.47) \\
Unicorn        & \underline{1.27(1.02)} & \underline{4.75(2.27)} & \underline{1.03(0.57)} & \underline{390.70(319.97)} & \underline{183.08(246.46)} & 2.70(1.23) & 4.55(1.43) & 1.20(0.46) & 528.44(450.53) & 253.57(327.96) \\
CROSS          & \textbf{1.05(0.91)} & \textbf{5.07(2.26)} & \textbf{1.04(0.61)} & \textbf{343.80(274.96)} & \textbf{152.44(214.65)} & \underline{0.51(0.13)} & \underline{8.10(0.34)} & \underline{1.38(0.52)} & \underline{303.39(162.93)} & \underline{47.56(62.27)} \\
\hline
\end{tabular}}
\end{table*}

\begin{table}[t]
\centering
\caption{Zero-shot Performance cross Hangzhou datasets.}
\label{tab:hangzhou_results}
\renewcommand{\arraystretch}{1.3}
\resizebox{\linewidth}{!}{
\begin{tabular}{c|ccccc}
\hline
\multirow{2}{*}{\textbf{Method}} 
& \multicolumn{5}{c}{\textbf{Real-World Dataset Evaluation}} \\ \cline{2-6}

& \begin{tabular}[c]{@{}c@{}}Queue Length $\downarrow$ \\ (veh)\end{tabular} 
& \begin{tabular}[c]{@{}c@{}}Speed $\uparrow$ \\ (m/s)\end{tabular} 
& \begin{tabular}[c]{@{}c@{}}Trip Comp. Rate $\uparrow$ \\ (veh/s)\end{tabular} 
& \begin{tabular}[c]{@{}c@{}}Trip Time $\downarrow$ \\ (s)\end{tabular} 
& \begin{tabular}[c]{@{}c@{}}Trip Delay $\downarrow$ \\ (s)\end{tabular} \\ 
\hline

\multicolumn{6}{c}{\textbf{$Hangzhou_{(1)}$ (Medium, Real-World)}} \\
\hline
Fixed-Time     & 0.95(0.47) & 5.23(1.55) & 0.65(0.38) & 575.47(530.17) & 226.95(409.00)  \\
Max-Pressure   & \textbf{0.06(0.03)} & \textbf{9.47(0.28)} & \textbf{0.76(0.37)} & \textbf{332.22(169.81)} & \textbf{12.45(13.45)}  \\
\hline
GESA           & 0.25(0.10) & 8.43(0.53) & 0.75(0.37) & 374.42(264.41) & 57.89(192.77)  \\
Unicorn        & 0.53(0.19) & 6.90(0.69) & 0.74(0.37) & 458.48(298.37) & 121.98(157.47)  \\
CROSS          & \underline{0.17(0.06)} & \underline{8.81(0.30)} & \underline{0.75(0.37)} & \underline{356.01(194.82)} & \underline{37.84(59.76)}  \\
\hline

\multicolumn{6}{c}{\textbf{$Hangzhou_{(2)}$ (Hard, Real-World)}} \\
\hline
Fixed-Time     & 1.28(0.97) & 5.57(1.59) & 0.84(0.53) & 507.14(441.15) & 199.56(350.60)  \\
Max-Pressure   & \textbf{0.16(0.16)} & \textbf{9.13(0.49)} & \textbf{1.16(0.66)} & \textbf{325.12(176.56)} & \textbf{20.91(33.76)}  \\
\hline
GESA           &  0.45(0.53) & \underline{8.49(1.06)} & 1.11(0.61) & \underline{352.93(213.39)} & 59.76(108.32)  \\
Unicorn        & 1.39(1.12) & 5.75(1.44) & 0.92(0.51)  & 504.53(356.04) & 196.96(243.60) \\
CROSS          & \underline{0.42(0.34)} & 8.33(0.62) & \underline{1.13(0.64)} & 356.02(213.52) &  \underline{56.09(99.62)} \\
\hline

\end{tabular}}
\end{table}

\subsection{Results and Analysis}

\subsubsection{Evaluation on Synthetic Datasets}
We first evaluate our CROSS on the synthetic traffic datasets with other compared baselines.
The detailed results are summarized on the left side of Table~\ref{tab:synthetic_results}.
Traditional methods perform reasonably in the easy dataset (e.g., Grid $4\times4$), but deteriorate significantly as traffic demand and dataset complexity increase, reflecting their limited adaptability.
GESA achieves the best performance in easy datasets, with the lowest queue length (0.04 veh) and trip delay (19.12 s) in Grid $4\times4$. 
However, its performance drops drastically in medium and hard datasets, indicating its insufficient generalization to asymmetric and complex traffic flows.
Unicorn improves representation through both GFE and ISR modules, offering more stable and better performance across datasets, but constrained by its one-size-fits-all design, it still lacks specialization and diversity, limiting generalization capacity.
 
In contrast, CROSS combines clustering with a Scenario-Adaptive MoE, dynamically activating experts for different traffic patterns.
This enables flexible representation learning, consistently outperforming all baselines. 
Although we do not explicitly model coordination across intersections, our PCC and MoE modules implicitly account for neighboring interactions through movement-level traffic features.
On the challenging Grid $5\times5$, CROSS achieves the lowest queue length (1.05 veh), shortest trip time (343.80 s), lowest trip delay (152.44 s), and highest speed (5.07 m/s) and trip completion rate (1.04 veh/s).
Notably, trip delay, trip time, and trip completion rate should be interpreted together, as these metrics only account for vehicles that successfully reach their destinations. 
Vehicles still en route or stuck in congestion at the end of the simulation are excluded, which can lead to an overly optimistic assessment of performance. 

\begin{figure}
    \centering
    \includegraphics[width=\linewidth]{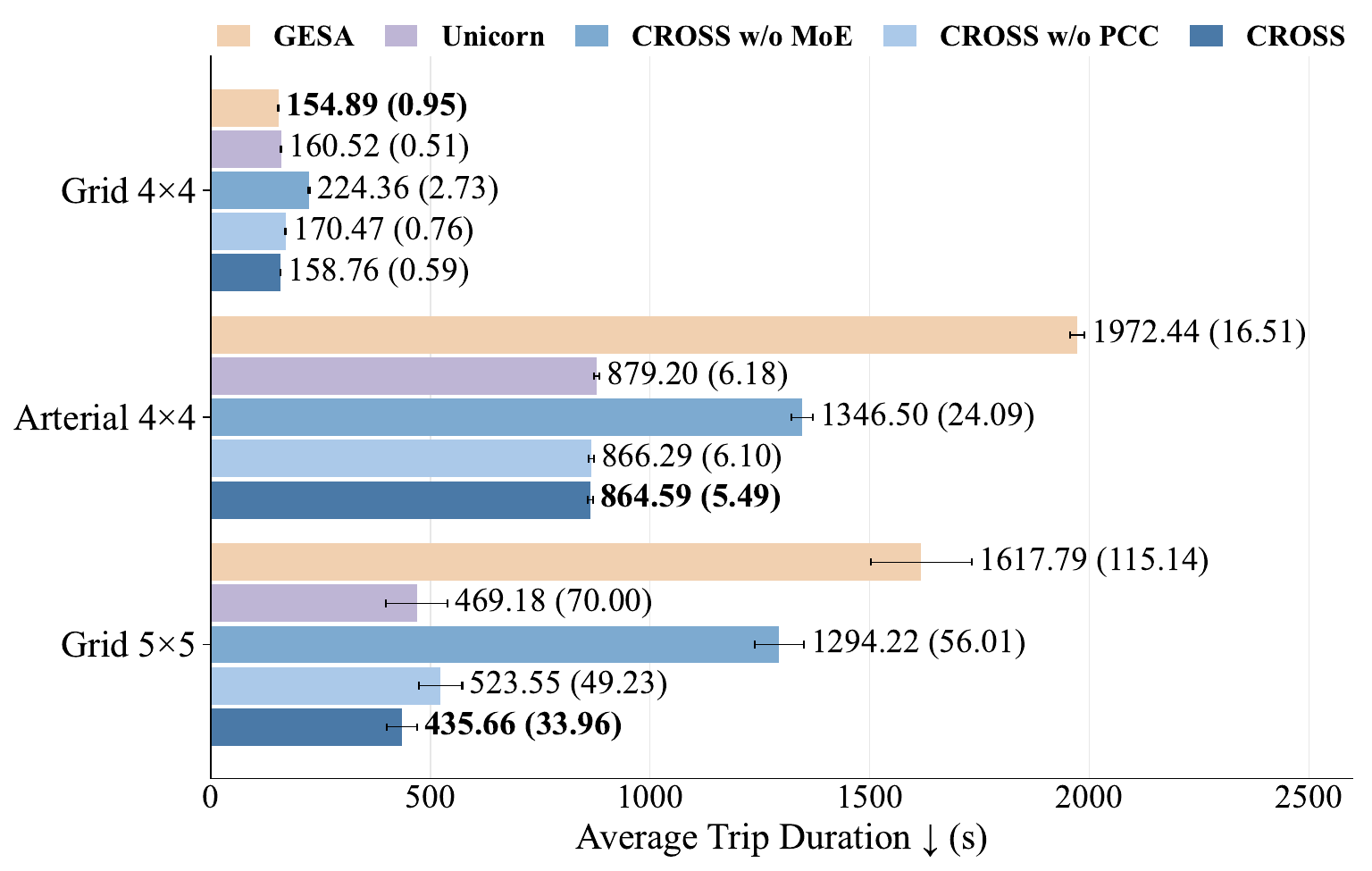}
    \caption{Average trip duration across three synthetic training datasets comparing CROSS with baseline methods (GESA, Unicorn) and ablations (CROSS w/o MoE, CROSS w/o PCC). Here $\downarrow$ denotes lower is better.}
    \label{fig:bar_chart}
\end{figure}

To obtain a fairer and more comprehensive evaluation, we adopt \textit{average trip duration}.
The corresponding results are shown in Fig.~\ref{fig:bar_chart}, where CROSS consistently achieves lower average trip duration than baselines and ablations.
These results once again validate the leading position of CROSS, demonstrating that our model can withstand challenges from all aspects, further highlighting its outstanding performance.

\subsubsection{Zero-shot Evaluation on Real-World Datasets}

To further assess the generalizability of CROSS, we conduct zero-shot evaluations on real-world datasets, with results reported on the right side of Table~\ref{tab:synthetic_results} and Table~\ref{tab:hangzhou_results}.
Max-Pressure achieves the strongest performance across real-world datasets. 
As a rule-based controller, it directly minimizes intersection pressure without relying on learned representations, which makes it naturally robust to domain shifts and helps maintain strong zero-shot performance.

In contrast, learning-based approaches learn from training data, which may lead to performance drops under domain shifts.
Although Unicorn offers strong adaptability across datasets, its one-size-fits-all design lacks the ability to extract universal patterns, causing it to overfit training characteristics and perform poorly in zero-shot evaluation.
GESA shows relatively stable behavior due to its unified structure mapping strategy that enhances cross-scenario consistency.
However, such a globally shared design may become suboptimal in heterogeneous traffic conditions, as it limits scenario-specific specialization.
Despite these challenges, CROSS significantly outperforms other learning-based baselines. 
The PCC module extracts universal and diverse representations, while the MoE enables adaptive expert activation for different traffic patterns, improving generalizability under domain shifts.

Overall, the zero-shot results reveal a practical trade-off: rule-based methods provide inherent robustness to distribution shifts, 
whereas CROSS minimizes this performance gap compared to other SOTA baselines, while preserving flexible representation learning and scenario-aware specialization.

\subsubsection{Ablation Study}
Fig.~\ref{fig:bar_chart} validates the effectiveness of both PCC and Scenario-Adaptive MoE modules.
Compared with the full CROSS model, the CROSS w/o PCC variant shows a clear increase in average trip duration, particularly in Grid $5\times5$ dataset with an approximate 20\% degradation. 
Its performance becomes close to Unicorn, indicating that explicit pattern modeling enhances representation capacity.
The CROSS w/o MoE variant performs significantly worse, further highlighting the critical role of the proposed MoE module in generating scenario-adaptive policies.

%%%%%%%%%%%%%%%%%%%%%%%%%%%%%%%%%%%%%%%%%%%%%%%%%%%%%%%%%%%%%%%%%%
\section{CONCLUSION}
\label{sec:conclusion}
In this paper, we propose CROSS, a novel MoE-based decentralized RL framework for generalizable large-scale ATSC via cross-scenario joint training.
To achieve this, we introduce a PCC module for abstracting generalized patterns, paired with a Scenario-Adaptive MoE for deriving flexible, scenario-specific control policies.
Specifically, PCC is designed to form a discriminative representation space that separates distinct traffic patterns.
Conditioned on these patterns, the Scenario-Adaptive MoE selectively activates the most suitable experts for scenario-specific control.
This synergistic integration offers a broader perspective on policy learning, which shifts the paradigm from monolithic control to adaptive specialization, thereby achieving a balance between universal generalizability and fine-grained representation capacity.
Although trained on synthetic data, CROSS achieves improved control performance and robust zero-shot transfer to real-world scenarios, surpassing existing methods and indicating its potential for practical ATSC deployment.

In future work, we will explore the applicability of the proposed PCC module and Scenario-Adaptive MoE module to other multi-agent and robotic systems.
By doing so, we hope to establish CROSS as a general paradigm to facilitate efficient and specialized strategy learning in environments with diverse agent roles and varying state and action spaces.

%%%%%%%%%%%%%%%%%%%%%%%%%%%%%%%%%%%%%%%%%%%%%%%%%%%%%%%%%%%%%%%%%
% \section*{ACKNOWLEDGMENTS}

%%%%%%%%%%%%%%%%%%%%%%%%%%%%%%%%%%%%%%%%%%%%%%%%%%%%%%%%%%%%%%%%%%

\bibliographystyle{IEEEtran}
\bibliography{root} 

\end{document}